\documentclass[sigconf,screen,nonacm]{acmart}

\setcopyright{rightsretained}
\acmYear{2027}

\usepackage{booktabs,tabularx,multirow,enumitem,tikz,pgfplots,array}
\pgfplotsset{compat=1.18}
\usetikzlibrary{arrows.meta,positioning}

\newcommand{\code}[1]{\texttt{#1}}
\newcommand{\itt}{ITT}
\newcommand{\crapi}{CR$_{\mathrm{api}}$}
\newcommand{\critt}{CR$_{\mathrm{itt}}$}
\newcommand{\spdf}{\ensuremath{S_{\mathrm{pdf}}}}
\newcommand{\docmut}{\textsc{DocMut}}

\hypersetup{urlcolor=black}

\begin{document}

\title{TeXFix-Bench: An Empirically Grounded Multi-Format Benchmark for
LLM-Based Document Source Repair}

\author{Prajwal S. Venkateshmurthy}
\orcid{0009-0004-8830-0725}
\affiliation{%
  \institution{Independent Researcher}
  \city{San Jose}
  \state{CA}
  \country{USA}
}
\email{prajwal@prajwal.me}

\begin{abstract}
Scientific and technical writing depends on markup sources that must
\emph{compile}: LaTeX, Typst, and Markdown pipelines fail when documents
contain missing delimiters, mismatched environments, broken imports, or package
conflicts. Existing document-repair evaluations inject faults with ad-hoc text
edits that lack an empirically derived fault model. We present
\emph{TeXFix-Bench}, a multi-format benchmark for LLM-based full-source
document repair grounded in a mined fault taxonomy. First, a Grounded-Theory
study of localized hard-crash LaTeX faults mined from TeX~Stack~Exchange,
GitHub commits, and package documentation (168 verified faults, dual open
coding at $\kappa{=}0.34$ with a documented granularity analysis) yields an
18-category taxonomy that we instantiate as \docmut{}: 48 AST-aware
operators across three formats (the taxonomy is mined for LaTeX, and Typst
and Markdown operators extend it by analogy). A three-model cross-benchmark shows \docmut{} faults are
5.6--9.2\,pp harder to repair than pattern-based mutations on the same seeds,
and a real-error case study (88 mined human crashes, 67.0\% repair success)
brackets both synthetic sets from below. Second, we construct 10{,}437 mutation
instances from 743 openly licensed seeds and evaluate seven contemporary LLMs
under a fixed zero-shot protocol with provider-pinned routing, collecting
48{,}651 attempts at a total inference cost of ${\approx}$\$200. A complete
6{,}613-instance $\times$ 7-model balanced matrix confirms all rankings. A
pinned engine gate yields a 27.5-point intention-to-treat compile spread
(56.7--84.2\%). Typst is markedly harder than LaTeX and Markdown. Third, a
restoration oracle over 28{,}129 compiling repairs shows that 13.6--18.5\%
of compiling repairs materially alter document text, and
that restoration rank diverges from compile rank: the model with the lowest
compile rate restores content best among its successes. Compile success alone
overstates repair quality. We release the taxonomy, \docmut{}, and all
campaign artifacts.
\end{abstract}



\maketitle

\section{Introduction}
\label{sec:introduction}

Markup sources dominate scientific and technical authoring: LaTeX for
papers and theses, Markdown for documentation and computational notebooks
(Jupyter, R~Markdown, Quarto), and Typst as a fast-growing newcomer.
When sources fail to compile or convert, authors face opaque diagnostics,
package/engine interactions, and multi-file project complexity. LLM-based
assistants are increasingly offered as a fix: paste the broken source, receive
a repaired document. How well this works, and for which formats, faults, and
models, has not been measured with the discipline used for code repair.

Automated program repair (APR) research has produced strong executable
benchmarks such as Defects4J~\cite{just2014defects4j},
ManyBugs~\cite{legoues2015manybugs}, and SWE-bench~\cite{jimenez2024swebench}.
Their contracts rest on unit tests or issue-reproduction environments. Document
markup has a different oracle: engines and converters (Tectonic/pdfTeX,
Typst, pandoc) decide whether a source is buildable. Fault patterns also differ
from typical programming bugs: unbalanced groups, environment mismatches,
missing packages, invalid set-rules, and shell-escape requirements.

Mutation-based evaluation solves the supply problem, because real
broken documents are scarce, poorly licensed, and multi-file. But it
introduces a scientific risk: \emph{unrealistic mutants}. If injected faults do
not resemble errors authors actually make, repair scores misstate field
performance. Code mutation addressed this by grounding operators in fault
taxonomies~\cite{offutt1996experimental,jia2011mutation}. Document repair has
lacked an analogous foundation. Prior document-repair benchmarks (including our
own v0.3 pilot~\cite{texfixv03}) injected faults with pattern-based text edits
that are cheap and reproducible but unvalidated against human fault
distributions.

\begin{figure*}[t]
\centering
\resizebox{\textwidth}{!}{%
\begin{tikzpicture}[
  font=\small,
  box/.style={draw=#1!70!black,fill=#1!12,rounded corners=3pt,
    minimum height=1.2cm,text width=2.5cm,align=center,line width=.6pt},
  arrow/.style={-{Stealth[length=2mm]},line width=.7pt,draw=black!50}
]
\node[box=teal] (a) {Mine human faults\\TeX.SE, GitHub, docs\\168 localized};
\node[box=blue,right=0.5cm of a] (b) {Taxonomy\\18 categories\\dual-coded};
\node[box=orange,right=0.5cm of b] (c) {\docmut{}\\48 operators\\3 formats};
\node[box=purple,right=0.5cm of c] (d) {Benchmark\\10{,}437 instances\\743 seeds};
\node[box=red,right=0.5cm of d] (e) {7-model campaign\\48{,}651 attempts\\provider-pinned};
\node[box=teal,right=0.5cm of e] (f) {Two oracles\\compile gate\\restoration \spdf};
\draw[arrow] (a)--(b);
\draw[arrow] (b)--(c);
\draw[arrow] (c)--(d);
\draw[arrow] (d)--(e);
\draw[arrow] (e)--(f);
\draw[arrow,dashed] (c.south) to[out=-30,in=-150]
  node[below,font=\scriptsize,align=center]{validation: 3-model cross-benchmark,\\real-error case study} (f.south);
\end{tikzpicture}}
\caption{End-to-end pipeline: mined faults ground the taxonomy, the taxonomy
grounds the operators, and the resulting benchmark is scored under both a
compile oracle and a content-restoration oracle.}
\label{fig:pipeline}
\end{figure*}
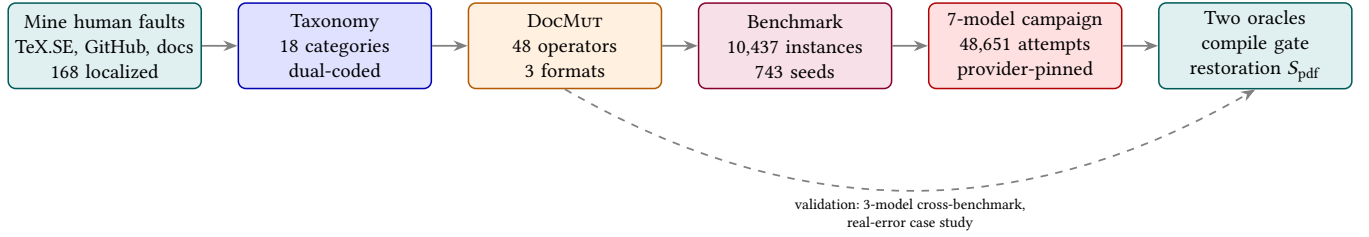

This paper closes that loop end to end (Figure~\ref{fig:pipeline}): we \emph{mine} real compilation
faults, \emph{derive} a taxonomy, \emph{instantiate} it as a mutation library,
\emph{validate} that the library produces harder and more realistic repair
tasks than pattern-based injection, and then \emph{evaluate} seven
contemporary LLMs at scale under both a compile oracle and a content
restoration oracle.

\paragraph{Contributions.}
\begin{enumerate}[leftmargin=*]
  \item \textbf{A mined fault taxonomy for LaTeX compilation crashes}
  (Section~\ref{sec:taxonomy}): a dual-sampled mining pipeline over TeX.SE,
  GitHub commits, and package documentation with hard-crash verification and
  explicit exclusion funnels. The result is 168 localized faults dual-coded into 18 axial categories, with an honest reliability analysis ($\kappa{=}0.34$ under
  free-form open coding, diagnosed as granularity asymmetry, plus a
  codebook-operationalizability check at $\kappa{=}0.64$).
  \item \textbf{\docmut{}}~\cite{docmut} (Section~\ref{sec:docmut}): a
  48-operator, multi-format mutation library (25 LaTeX + 15 Typst + 8
  Markdown) with hard/soft tracks, difficulty tiers, deterministic PRNG
  control, and a render-diff equivalence guard (controlled F1${=}0.937$).
  \item \textbf{Operator validation} (Section~\ref{sec:validation}): a
  three-model cross-benchmark on matched seeds shows \docmut{} faults are
  \textbf{5.6--9.2\,pp harder to repair} than pattern-based mutations for
  every model family tested. A real-error case study (88 mined human hard
  crashes, 67.0\% repair success) shows human faults are harder still.
  \item \textbf{TeXFix-Bench v0.4} (Section~\ref{sec:benchmark}): 10{,}437
  unique hard instances from 743 openly licensed seeds drawn from five
  public source families (template packs, Overleaf, CTAN, GitHub, Typst
  Universe) plus pandoc-derived Markdown, engine-gated under pinned
  Tectonic/Typst/pandoc.
  \item \textbf{A seven-model hosted evaluation} (Sections~\ref{sec:method},
  \ref{sec:results}): 48{,}651 attempts at a transparently accounted
  ${\approx}$\$200 total inference cost with
  provider-pinned routing (\code{allow\_fallbacks=false}) and a per-request
  provider ledger. Primary claims verified on a \textbf{complete
  6{,}613-instance $\times$ 7-model balanced matrix}.
  \item \textbf{A two-oracle result} (Section~\ref{sec:restoration}): beyond a
  27.5-point \itt{} compile spread and a clear Typst difficulty gap, a
  restoration oracle over 28{,}129 compiling repairs shows 13.6--18.5\% of
  compiling repairs materially alter document text, with restoration rank
  diverging from compile rank. Compile-only leaderboards overstate repair
  quality.
\end{enumerate}

\paragraph{Scope.}
We evaluate \emph{full-source} repair under a fixed harness. We do not claim a
universal model ranking, visual/layout PDF equivalence, or ecological coverage
of all real authoring errors. The mined taxonomy targets \emph{localized hard
crashes} under a fixed engine. Multi-hunk reconfigurations are an explicit
out-of-scope class (Section~\ref{sec:taxonomy}).

\paragraph{Paper roadmap.}
Section~\ref{sec:related} positions related work.
Sections~\ref{sec:taxonomy}--\ref{sec:validation} derive and validate the
fault model. Sections~\ref{sec:benchmark}--\ref{sec:method} describe the
benchmark and campaign. Section~\ref{sec:results} answers the research
questions. Sections~\ref{sec:discussion}--\ref{sec:threats} discuss
implications and threats.

\section{Related Work}
\label{sec:related}

\subsection{APR benchmarks}
Defects4J, ManyBugs/IntroClass, and SWE-bench fix task contracts around tests
or GitHub issues~\cite{just2014defects4j,legoues2015manybugs,jimenez2024swebench}.
Smith et al.\ warn that test-passing patches can overfit~\cite{smith2015overfitting}.
Monperrus surveys the broader APR literature~\cite{monperrus2018apr}. We adopt
the same caution for compile oracles. A minimal stub can ``pass''
compilation, so we address this directly with a restoration oracle
(Section~\ref{sec:restoration}).

\subsection{Mutation testing}
Surveys by Jia and Harman and by Papadakis et al.\ frame operator design,
equivalent mutants, and determinism~\cite{jia2011mutation,papadakis2019mutation}.
Offutt et al.\ established that operator selection should be grounded in
empirical sufficiency~\cite{offutt1996experimental}. Code mutators (PIT,
Stryker, Universal Mutator~\cite{universalmutator}) target application
languages. Markup constructs such as braces, environments, packages, and
engine flags have no direct analogue in those toolkits, which motivates
deriving a domain taxonomy \emph{before} operator design.

\subsection{LLM-based code repair}
DeepFix and diagnostic-feedback repair address student
programs~\cite{gupta2017deepfix,yasunaga2020graph}. Conversational APR systems
show strong results on classic bug datasets~\cite{xia2023alpharepair}. These
works motivate LLM repair but use different oracles and artifacts.

\subsection{Document-adjacent systems}
EqFix repairs individual equations~\cite{zhu2022eqfix}. TexOCR reconstructs
LaTeX from page images~\cite{wang2026texocr}. Editor assistants target writing
workflows rather than compile oracles~\cite{wen2024overleafcopilot,hou2026paperdebugger}.
Tan and Rigger study cross-engine inconsistencies among \emph{successfully}
typeset documents~\cite{tan2024tex}. Our setting is the dual: sources that do
not produce output at all. We pin engines accordingly.

\subsection{Qualitative coding in empirical SE}
Our dual open coding follows Grounded Theory practice and thematic analysis
guidance~\cite{braun2006thematic}. We report Cohen's $\kappa$ with Landis and
Koch bands~\cite{landis1977kappa} and discuss how free-form open coding
depresses $\kappa$ relative to fixed-codebook designs~\cite{viera2005kappa}.

\subsection{Positioning}
\emph{To our knowledge}, TeXFix-Bench is the first multi-format
(LaTeX, Typst, Markdown) engine-gated repair benchmark whose mutation
operators are validated against a mined human fault taxonomy, and the first to
report compile and content-restoration oracles jointly at this scale.

\section{An Empirically Grounded Fault Taxonomy}
\label{sec:taxonomy}

We target \emph{localized hard crashes}: single-site (or tightly coupled)
faults that produce non-zero engine exit codes under Tectonic. We deliberately
exclude multi-hunk reconfigurations, warning-only diagnostics, and multi-file
asset puzzles that confound mutation oracles.

\subsection{Mining pipeline}
We mine three \emph{human} sources for taxonomy derivation and keep a fourth
source (AI repair failures) for difficulty analysis only.

\paragraph{TeX Stack Exchange (dual sampling).}
From the TeX.SE data dump we select questions tagged \code{errors} or
\code{compilation-\allowbreak error} with accepted answers ($N{=}2{,}655$ universe):
\textbf{Popular} (top 500 by views) and \textbf{Random} (500 seeded draws,
zero overlap). Dual sampling reduces the risk that the taxonomy reflects only
highly upvoted threads~\cite{braun2006thematic}.

\paragraph{GitHub commits with tangled-commit isolation.}
Four search queries yield 1{,}662 unique commits across 1{,}122 repositories
(median patch size 337 lines). A raw ``fix + \code{.tex}'' commit is almost
never a pure compilation repair, so we apply a three-stage high-precision
funnel: (1)~exactly one \code{.tex} file, under 20 non-context lines, and a
message matching fix/compile/error/broken, leaving 71 survivors,
(2)~retain only fault-fix lines (delimiters, packages, commands, braces), and
(3)~manual verification on a random 50 (heuristic prelabel accuracy 0.94).
The 20-line threshold sits on the commit-size CDF: only 13.3\% of commits
are that small, and raising it to 30 lines adds 2.8\,pp of candidates while
admitting prose-mixed hunks.
The 95.7\% exclusion rate is a deliberate precision bias: multi-file and large
commits systematically confound localization and are the wrong seed material
for single-operator mutation design.

\paragraph{Package documentation.}
For 13 widely used packages we extract 30 hard-crash troubleshooting patterns
from manuals/CTAN (29 verified hard-crash). These seed missing-package,
option-clash, and engine-requirement faults that forums under-sample.

\paragraph{Hard-crash verification and localization.}
Every candidate is compiled with Tectonic. Only non-zero exits survive
(\textbf{237} verified crashes, 86.2\% survival among reconstructions). We
then classify on the \emph{post-fix} (compiling) version: diff
broken$\leftrightarrow$fixed, map changes to coarse AST node types, keep
single-node changes. Result: \textbf{168 localized} (70.9\%), and \textbf{69
reconfigurations} excluded, a distinct multi-concern class (babel option +
package changes, biblatex/preamble rewrites) that violates the
single-operator contract and is explicit future work, not discarded evidence.
By source: TeX.SE 143, GitHub 14, package docs 11.

\subsection{Dual coding and reliability}
\label{sec:kappa}
Annotator~1 (author) and Annotator~2 (independent) labeled all 168 faults with
free-form open codes and \emph{no shared codebook}. Open strings were then
mapped to 18 axial categories via documented synonym rules (coding handbook
released).

\paragraph{Reliability.}
On axial labels after synonym merge, Cohen's $\kappa{=}\mathbf{0.34}$
($p_o{=}0.41$, $p_e{=}0.11$), \emph{fair} under Landis and
Koch~\cite{landis1977kappa}. The primary cause is open-code granularity
asymmetry: Annotator~1 produced 22 consolidated labels, while Annotator~2
produced 120 near-instance-specific phrases. Many-to-one synonym collapse cannot
recover agreements that never existed at matching granularity. Residual
disagreements (e.g., ``undefined control sequence'' symptom vs.\ missing
package root cause) were resolved with diagnostic-priority and specificity
rules (audit trail released). Fair $\kappa$ under free-form open coding with
asymmetric aggregation is expected~\cite{braun2006thematic,viera2005kappa}.
We report it as a finding about coding methodology and treat a second-round
closed-codebook human annotation as future work.

\paragraph{Codebook operationalizability check.}
As a complementary check, \emph{not} a second inter-rater measurement, we
implemented the 18 categories plus the resolution rules as a deterministic
classifier over compiler diagnostics and fault-site features. It reproduces
Annotator~1's axial labels at $\kappa{=}0.64$ ($p_o{=}0.70$): the categories
are well-defined enough to be assigned mechanically from diagnostics at
substantial agreement with a human coder. Because the classifier encodes the
resolution rules, this does not substitute for human--human reliability.

\subsection{Taxonomy and frequencies}
Figure~\ref{fig:freq} shows the resolved distribution. Top categories among
the 168 resolved faults: UndefinedControlSequence
(18.5\%, Wilson 95\% CI $[13.3,25.0]$), OtherFatalCompile (10.1\%),
InputEncodingFault (9.5\%), BraceGroupFault and MathModeFault (8.9\% each),
GraphicsIncludeFault (7.1\%).

\begin{figure}[t]
\centering
\begin{tikzpicture}
\begin{axis}[
  ybar,
  bar width=3.6pt,
  width=\columnwidth,
  height=4.2cm,
  ymin=0, ymax=35,
  ylabel={Faults ($N{=}168$)},
  symbolic x coords={UCS,Other,Enc,Brace,Math,Gfx,Pkg,Tab,Doc,Font,EnvU,EnvM,File,Shell,Babel,Ams,Pream,Clash},
  xtick=data,
  x tick label style={rotate=55,anchor=east,font=\tiny},
  tick label style={font=\tiny},
  ylabel style={font=\scriptsize},
  enlarge x limits=0.04,
  nodes near coords,
  every node near coord/.append style={font=\tiny},
]
\addplot[fill=blue!55,draw=blue!70!black] coordinates {
  (UCS,31) (Other,17) (Enc,16) (Brace,15) (Math,15) (Gfx,12) (Pkg,9)
  (Tab,8) (Doc,8) (Font,7) (EnvU,6) (EnvM,6) (File,6) (Shell,4)
  (Babel,3) (Ams,2) (Pream,2) (Clash,1)
};
\end{axis}
\end{tikzpicture}
\caption{Frequency of resolved axial categories among the 168 localized
hard crashes. UCS = UndefinedControlSequence, Other = OtherFatalCompile,
Enc = InputEncoding, Brace = BraceGroup, Math = MathMode, Gfx =
GraphicsInclude, Pkg = MissingRequiredPackage, Tab = TableAlignment, Doc =
DocumentBoundary, Font = FontEngine, EnvU = UndefinedEnvironment, EnvM =
EnvironmentMismatch, File = MissingExternalFile, Shell = ShellEscape,
Babel = BabelLanguage, Ams = AmsmathStructure, Pream = PreambleMisuse,
Clash = PackageOptionClash.}
\label{fig:freq}
\end{figure}
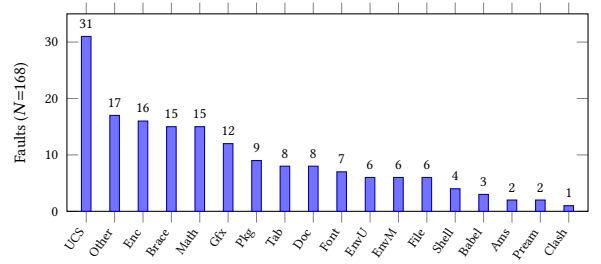 Of 18 nonempty categories, 13 (72.2\%) are
targeted by at least one \docmut{} LaTeX operator, and \textbf{instance
coverage} is 76.2\% (CI $[69.2,82.0]$). Uncovered categories
(InputEncodingFault, MissingExternalFile, BabelLanguageFault,
AmsmathStructureFault) are first-class backlog items.

\section{The DocMut Mutation Library}
\label{sec:docmut}

\docmut{} v0.2.0~\cite{docmut} instantiates the taxonomy as 48 operators: 25
LaTeX, 15 Typst, and 8 Markdown (Table~\ref{tab:ops-counts}). Operators are
implemented against format-specific structure (not blind regex) and labeled
with \emph{tier} and \emph{hard/soft track}. Hard operators are retained only
when the golden compiles and the broken source fails under the designated
engine. Soft operators target diagnostic/semantic faults that may still
compile. Site selection uses a deterministic PRNG. The mined taxonomy is
LaTeX-only. Typst/Markdown operators share the adequacy framework and are
designed by analogy (a scoped external-validity note,
Section~\ref{sec:threats}).

\begin{table}[t]
\caption{DocMut operator inventory.}
\label{tab:ops-counts}
\begin{tabular}{@{}lrrr@{}}
\toprule
Format & Operators & Hard track & Soft track \\
\midrule
LaTeX & 25 & 16 & 9 \\
Typst & 15 & 12 & 3 \\
Markdown & 8 & 0 & 8 \\
\midrule
Total & 48 & 28 & 20 \\
\bottomrule
\end{tabular}
\end{table}

\paragraph{Example.}
A Tier-1 \code{TEX-BRC-DRP} (BraceDrop) mutation removes one closing brace
at a PRNG-selected site, leaving a source that Tectonic rejects with a
runaway-argument error:
\begin{quote}
\small
golden: \code{(\textbackslash vec\{u\}-\textbackslash vec\{w\})}\\
broken: \code{(\textbackslash vec\{u\}-\textbackslash vec\{w)}
\end{quote}
A Tier-3 \code{TEX-SHL-ESC} mutation instead injects a \code{minted}
dependency whose repair requires an engine-flag or package substitution
rather than a local token edit, spanning the two ends of the difficulty
range.

\paragraph{Difficulty tiers.}
\begin{description}[leftmargin=*,style=nextline]
  \item[Tier 1 (surface/structural)] Delimiters, unclosed environments/blocks,
  missing class or import declarations.
  \item[Tier 2 (semantic/structure)] Wrong commands or types, argument drops,
  dictionary keys, soft environment swaps.
  \item[Tier 3 (realistic author patterns)] Shell-escape requirements,
  glossary/font/package-order issues, HTML/image mistakes.
\end{description}

\paragraph{Taxonomy-driven refinement.}
Pilot operators existed before the formal Grounded-Theory study. The study
validates and refines them: (1)~BraceDrop split into runaway vs.\
extra-brace subpatterns, (2)~PackageDrop refined into missing-package vs.\
option-clash hard patterns (excluding warning-only hyperref load-order
styles), (3)~soft FontSwap deprecated for hard-crash evaluation.

\paragraph{Render-diff equivalence guard.}
Equivalent mutants must be excluded from hard repair sets. We compile golden
and mutant under Tectonic, extract text with a fixed \code{pdftotext}
bounding box, normalize (form-feed strip, whitespace collapse, no
lowercasing), and compare. On 120 controlled pairs (true equivalents such as identity, whitespace, and comment edits, and true non-equivalents such as body text, math, and preamble edits), precision is 0.88, recall 1.0, \textbf{F1${=}0.937$},
over-normalization rate 0. Structure and math scopes reach F1${=}1.0$.
Preamble edits reach F1${=}0.85$ (style changes can leave body text unchanged).

\section{Operator Validation}
\label{sec:validation}

Does taxonomy-guided injection actually change measured repair difficulty? We
answer with a controlled cross-benchmark, a per-operator analysis, and a
real-error case study.

\subsection{Cross-benchmark: pattern-based vs.\ DocMut}
\label{sec:crossbench}
From the same seed pool we build two equal-sized hard-fail mutation sets
($n{=}391$ each): \textbf{Set~A}, pattern-based mutations (delete brace /
misspell command / drop package line), extending the injection style of
earlier document-repair benchmarks. \textbf{Set~B}, \docmut{} hard-track
operators on matched seeds. Both sets require broken sources that fail to
compile. Empty API responses count as missing, not successful repairs. We
repair both sets with three models from distinct families
(Table~\ref{tab:cross-multi}).

\begin{table}[t]
\caption{Cross-benchmark: compile success (exact match) by model, $n{=}391$
per set. Gap $=$ Set~A $-$ Set~B compile success.}
\label{tab:cross-multi}
\centering
\begin{tabular}{@{}lccc@{}}
\toprule
Model & Set A (pattern) & Set B (DocMut) & Gap (pp) \\
\midrule
DeepSeek~V4~Pro & 90.5\% (48.1\%) & 81.3\% (33.5\%) & \textbf{9.2} \\
Grok-4.3 & 91.6\% (26.1\%) & 85.9\% (23.3\%) & \textbf{5.6} \\
Codestral-2508 & 89.0\% (14.8\%) & 80.1\% (10.5\%) & \textbf{9.0} \\
\midrule
\multicolumn{4}{@{}l}{\footnotesize Compile counts of 391: DeepSeek 354/318,
Grok 358/336, Codestral 348/313.} \\
\bottomrule
\end{tabular}
\end{table}

The direction replicates in all three families: \docmut{} faults are
\textbf{5.6--9.2\,pp harder} under compile success and 2.8--14.6\,pp harder
under exact match, with nearly identical API completion ($\ge$97.7\%) and
cost. The strongest model (Grok) shows the smallest gap, but no model finds
\docmut{} faults easier. All three gaps are statistically significant under
exact McNemar tests on the seed-paired outcomes: DeepSeek $b{=}60$/$c{=}24$
with $p{<}0.001$, Grok $b{=}42$/$c{=}20$ with $p{=}0.007$, and Codestral
$b{=}65$/$c{=}30$ with $p{<}0.001$, where $b$ counts seeds repaired only in
Set~A and $c$ only in Set~B. \emph{How} faults are injected changes measured
repair difficulty by up to ten points under a fixed model and seed pool.

\subsection{Per-operator difficulty (Set B, DeepSeek)}
High compile success: BraceStray 95.6\%, EnvRename 93.1\%, ItemMisplaced
91.3\%. Low: ShellEscapeReq \textbf{47.2\%}, MathDisplay 50.0\% (small $n$),
DocumentClassDrop 65.4\%. Exact-match rates sit far below compile rates
throughout (e.g., PackageDrop 77.8\% compile but 5.6\% exact): models often
compile via non-minimal rewrites, inventing substitute packages or deleting
dependent body code rather than restoring the original line. This gap
motivates the restoration oracle of Section~\ref{sec:restoration}.

\subsection{Real-error case study}
\label{sec:realerror}
For ecological validity we also repaired mined human hard crashes
(DeepSeek~V4~Pro). From the 168 localized faults we stratified 100 cases.
Excluding custom-class documents and one non-failing sandbox case leaves
\textbf{88} evaluated cases (dummy assets provisioned, \code{.tex}
unmodified). Compile success is \textbf{67.0\%} (59/88), below \docmut{}
Set~B (81.3\%) and far below pattern-based Set~A (90.5\%)
(Table~\ref{tab:threeway}). In-sample: math delimiters 9/9 and brace balance
5/5, but input encoding 3/8, undefined control sequences 4/10, missing files
1/4. \textbf{Case study only}, not a population estimate, but the ordering
(pattern $>$ \docmut{} $>$ real) indicates taxonomy-guided injection moves
synthetic evaluation \emph{toward} real difficulty, not past it.

\begin{table}[t]
\caption{Same model, three fault sources (DeepSeek~V4~Pro).}
\label{tab:threeway}
\centering
\begin{tabular}{@{}lrr@{}}
\toprule
Fault source & $n$ & Compile success \\
\midrule
Pattern-based mutations (Set A) & 391 & 90.5\% \\
DocMut mutations (Set B) & 391 & 81.3\% \\
Real human hard crashes & 88 & \textbf{67.0\%} \\
\bottomrule
\end{tabular}
\end{table}

\section{Benchmark Construction}
\label{sec:benchmark}

\subsection{Dataset construction}
Seeds are real, openly licensed documents (no LLM-authored templates):
template packs (CC0), Overleaf Gallery (CC BY / CC0), CTAN examples (LPPL),
GitHub LaTeX/Typst repositories (MIT / Apache / CC0), Typst Universe
packages, and pandoc-converted Markdown derived from simple LaTeX. After
filtering and SHA-256 deduplication of broken sources, the stratified freeze
contains \textbf{10{,}437} instances from \textbf{743} unique seeds
(Table~\ref{tab:corpus}). Instance-level licenses: MIT (4{,}543), CC-BY-4.0
(2{,}575), CC0-1.0 (1{,}583), LPPL-1.3c (1{,}045), Apache-2.0 (508), MIT-0
(183).

\begin{table}[t]
\caption{TeXFix-Bench v0.4 corpus (frozen evaluation set).}
\label{tab:corpus}
\begin{tabular}{@{}lrrr@{}}
\toprule
Format & Instances & Unique seeds & Engine \\
\midrule
LaTeX & 6{,}568 & 395 & Tectonic \\
Typst & 3{,}263 & 188 & Typst \\
Markdown & 606 & 160 & pandoc \\
\midrule
Total & 10{,}437 & 743 & --- \\
\bottomrule
\end{tabular}
\end{table}

\subsection{Instance generation}
Each instance records \code{id}, format, \code{operator\_code}, tier, track,
\code{golden\_source}, \code{broken\_source}, content hashes, mutation site
metadata, and \code{compile\_engine}. Construction uses a fixed salt
(\code{20260802}) for deterministic PRNG site selection. The \docmut{} catalog defines 48 operators. The frozen benchmark
instantiates 39 of them with nonzero instance support after gate
filtering. The other nine produced zero retained mutants under the
construction budget (no eligible sites, gate failures, or inadequate soft
mutations), and one ({\code{TEX-MTH-OPS}}) has single-instance support. The
engine-gate ledger therefore covers \textbf{38} operators
(Appendix~\ref{app:ops}). Tier counts are 6{,}364 (Tier~1), 2{,}620 (Tier~2), and 1{,}453 (Tier~3).

\subsection{Engine-gate oracle}
Designated engines match construction: Tectonic 0.17.0 (LaTeX), Typst
0.15.1 (Typst), pandoc 3.10.1 (Markdown, HTML conversion primary). Each
compile uses a fresh temporary directory and a 30\,s wall-clock timeout.
Controls on campaign-touched instances ($N{=}7{,}007$): golden compile rate
\textbf{100\%}. Broken residual compile rate \textbf{5.7\%}, concentrated in
soft-track and Markdown operators retained for structural change even when
conversion still succeeds. Equivalent-mutant detection compares SHA-256 of
the candidate (after light fence unwrap) to the golden hash.

\section{Evaluation Methodology}
\label{sec:method}

\subsection{Models and providers}
Table~\ref{tab:models} lists the seven models accessed through
OpenRouter~\cite{openrouter} with \code{allow\_fallbacks=false} (no silent
provider substitution). Because hosted delivery is part of our results, we
release a per-request provider ledger. Table~\ref{tab:models} summarizes it.
Four models were served $\ge$99.9\% by their first-party providers, so the
delivery failures reported below are properties of those serving stacks, not
marketplace routing noise.

\begin{table}[t]
\caption{Model panel with upstream serving providers (from the per-request
audit ledger, \code{allow\_fallbacks=false}).}
\label{tab:models}
\scriptsize
\begin{tabular}{@{}llp{0.45\columnwidth}@{}}
\toprule
Route & Type & Serving provider(s) \\
\midrule
\code{x-ai/grok-4.3} & closed & xAI 100\% \\
\code{deepseek/deepseek-v4-pro} & open & DeepSeek 100\% \\
\code{deepseek/deepseek-v4-flash} & open & Baidu 58\%, StreamLake 37\%, other 5\% \\
\code{z-ai/glm-5.2} & open & Decart 60\%, Novita 21\%, StreamLake 18\% \\
\code{qwen/qwen3.7-max} & open & Alibaba 100\% \\
\code{mistralai/codestral-2508} & open & Mistral 100\% \\
\code{meta-llama/llama-4-maverick} & open & DeepInfra 95\%, DigitalOcean 5\% \\
\bottomrule
\end{tabular}
\end{table}

\subsection{Prompt design}
We use a single zero-shot prompt for all models, temperature $0$,
\code{max\_tokens}${=}4096$ (ablated at 16{,}384 in
Section~\ref{sec:results}), no tools, no streaming. The system message
(format substituted) is:
\begin{quote}
\small
\emph{You are a document repair expert. You are given a \{FORMAT\} document that
FAILS to compile to PDF. Return ONLY the fully corrected \{FORMAT\} source code.
Do not add commentary, do not use code fences, do not add markdown formatting.
Preserve all content; change only what is needed to make the document compile
cleanly.}
\end{quote}
The user message is the raw \code{broken\_source}.

\subsection{API campaign and the balanced matrix}
\label{sec:campaign}
The planned matrix is $10{,}437\times 7=73{,}059$ requests (one repetition per
pair). The campaign ran with checkpointed scheduling at concurrency 10, retrying
only transient 5xx/429/timeout responses. All outcomes are retained under
intention-to-treat. Evaluation recorded \textbf{48{,}651} attempts at a
total provider cost of ${\approx}$\$200 (per-model cost breakdown in the
artifact), full replication of the campaign is therefore accessible to any
modestly resourced group. The schedule prioritized complete model coverage
per instance over complete instance coverage: every model covered
7{,}006--7{,}007 instances except DeepSeek-Flash (6{,}613). The intersection
yields a \textbf{complete balanced matrix of 6{,}613 instances $\times$ 7
models} (46{,}291 attempts, 95.1\% of recorded volume). All primary claims
are stated on this balanced matrix, with full-\itt{} numbers in the artifact.
Remaining schedule cells are unfinished, not silently dropped successes.

\subsection{Metrics}
\begin{description}[leftmargin=*,style=nextline]
  \item[API delivery] Fraction of attempts that return a usable candidate
  (ledger status \code{ok}).
  \item[\crapi{}] Compile success among API-ok attempts (conditional skill).
  \item[\critt{}] Compile success among \emph{all} attempts
  (deployment-facing rate that embeds delivery).
  \item[Equivalent-mutant rate] Fraction of API-ok candidates whose SHA-256
  equals the golden (exact reverts).
  \item[Restoration \spdf{}] Token-LCS Dice between candidate and golden
  \code{pdftotext} output for compiling non-revert candidates
  (Section~\ref{sec:restoration}).
\end{description}
We report rates with explicit denominators, and Wilson 95\% CIs on \critt{}.
\crapi{} is the model-skill metric, \critt{} additionally reflects each
serving stack. We do not treat either alone as a universal ranking.

\section{Results: Compile Oracle}
\label{sec:results}

Unless noted, numbers are from the balanced matrix
(Section~\ref{sec:campaign}). Full-set \itt{} rates differ by at most
0.6\,pp and preserve all orderings.

\subsection{RQ1: How do contemporary LLMs perform?}
\begin{figure}[t]
\centering
\begin{tikzpicture}
\begin{axis}[
  ybar,
  bar width=7pt,
  width=\columnwidth,
  height=4.6cm,
  ymin=0, ymax=100,
  ylabel={Rate (\%)},
  symbolic x coords={Grok,Codestral,Maverick,DS-Flash,DS-Pro,GLM,Qwen},
  xtick=data,
  x tick label style={rotate=25,anchor=east,font=\scriptsize},
  tick label style={font=\scriptsize},
  ylabel style={font=\scriptsize},
  legend style={at={(0.02,0.02)},anchor=south west,draw=none,fill=none,font=\scriptsize},
  ymajorgrids=true,
  grid style={gray!20},
]
\addplot[fill=blue!55,draw=blue!70!black] coordinates {
  (Grok,84.2) (Codestral,77.5) (Maverick,77.4) (DS-Flash,71.0)
  (DS-Pro,65.8) (GLM,64.9) (Qwen,56.7)
};
\addplot[fill=teal!60,draw=teal!80!black] coordinates {
  (Grok,84.2) (Codestral,77.6) (Maverick,77.6) (DS-Flash,82.8)
  (DS-Pro,86.9) (GLM,93.8) (Qwen,94.4)
};
\legend{\critt{},\crapi{}}
\end{axis}
\end{tikzpicture}
\caption{Deployment-facing \critt{} vs.\ conditional \crapi{} on the
balanced matrix. High conditional skill with low delivery (Qwen, GLM,
DS-Pro) yields weak end-to-end rates.}
\label{fig:models}
\end{figure}
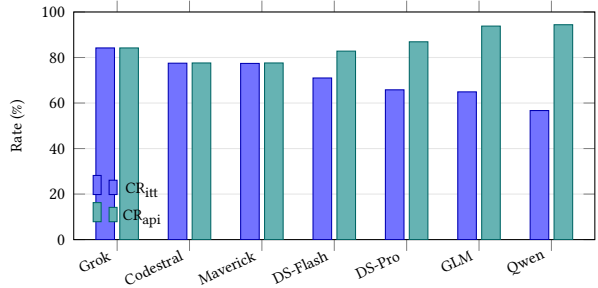

Table~\ref{tab:overall} and Figure~\ref{fig:models} summarize primary outcomes. Under \critt{},
Grok-4.3 leads at \textbf{84.2\%} with perfect delivery, while Qwen3.7-max is
lowest at \textbf{56.7\%} despite the highest conditional \crapi{} (94.4\%)
among delivering attempts, a 27.5-point spread. Codestral and Llama-4
Maverick combine $\ge$99.7\% delivery with mid-70\% \itt{}. GLM and
DeepSeek-Pro show low delivery with high conditional compile rates. Because
Qwen and DeepSeek-Pro were served 100\% by their first-party providers
(Table~\ref{tab:models}), their delivery deficits are serving-stack
properties, not router artifacts.

\begin{table*}[t]
\caption{Overall results by model on the balanced matrix.
Delivery = API-ok$/N$, \crapi{} = compile among API-ok, \critt{} = compile
among all attempts. 95\% Wilson CIs on \critt{}.}
\label{tab:overall}
\scriptsize
\begin{tabular}{@{}lrrrrrrl@{}}
\toprule
Model & $N$ & API-ok & Compile & Delivery & \crapi{} & \critt{} & 95\% CI \\
\midrule
x-ai/grok-4.3 & 6613 & 6613 & 5565 & 100.0\% & 84.2\% & \textbf{84.2\%} & $[83.3,85.0]$ \\
mistralai/codestral-2508 & 6613 & 6602 & 5122 & 99.8\% & 77.6\% & 77.5\% & $[76.4,78.4]$ \\
meta-llama/llama-4-maverick & 6613 & 6595 & 5118 & 99.7\% & 77.6\% & 77.4\% & $[76.4,78.4]$ \\
deepseek/deepseek-v4-flash & 6613 & 5672 & 4695 & 85.8\% & 82.8\% & 71.0\% & $[69.9,72.1]$ \\
deepseek/deepseek-v4-pro & 6613 & 5011 & 4352 & 75.8\% & 86.9\% & 65.8\% & $[64.7,66.9]$ \\
z-ai/glm-5.2 & 6613 & 4579 & 4295 & 69.2\% & 93.8\% & 64.9\% & $[63.8,66.1]$ \\
qwen/qwen3.7-max & 6613 & 3972 & 3749 & 60.1\% & \textbf{94.4\%} & 56.7\% & $[55.5,57.9]$ \\
\bottomrule
\end{tabular}
\end{table*}

\subsection{RQ2: Which document formats are hardest?}
On the full recorded set, Markdown is easiest under \critt{}
(\textbf{90.2\%}, $N{=}2{,}789$), then LaTeX (\textbf{74.2\%},
$N{=}30{,}814$), then Typst (\textbf{60.3\%}, $N{=}15{,}048$). Conditional on
API-ok, Markdown is near ceiling (99.9\%), LaTeX 85.5\%, Typst 77.7\%. The
Typst gap appears across nearly all models (per-model Typst \critt{} spans
42.8--86.1\%) and is consistent with newer syntax and less public training
mass, though we do not observe training corpora directly.

\subsection{RQ3: Does difficulty scale with tier?}
The tier gradient is modest under \critt{} (T1 71.5\%, T2 71.4\%, T3
66.9\%) and clearer conditionally (\crapi{} 86.0/84.6/76.1\%). Tier ordering
also varies across formats (Markdown Tier~3 is its \emph{easiest} cell).
For cross-format tier ordering, RQ3 is therefore largely a null result. We
treat tiers as per-format design labels rather than a universal difficulty
scale, and the per-operator view (RQ4) carries more signal.

\subsection{RQ4: Which operators are most challenging?}
Among operators with $\ge$20 API-ok attempts, the hardest are Typst import
drop (\code{TYP-IMP-DRP}, 17.5\% \crapi{}), LaTeX shell-escape requirement
(\code{TEX-SHL-ESC}, 31.8\%), and Typst unclosed math (48.4\%). Structural
import/package faults remain harder than local token typos, matching the
taxonomy's prediction that nonlocal dependency restoration is the hard core.
Operators at 100\% \crapi{} are mostly small-$n$ Markdown soft-track cells
and should not be read as ceilings.

\subsection{RQ5: What are the dominant failure modes?}
\paragraph{API-level.}
Of 48{,}651 recorded attempts, 4{,}352 returned empty, 2{,}258 hit the
completion cap, 1{,}051 failed in transport, and 58 hit residual rate
limits. Qwen and GLM concentrate
empty and length failures. Grok, Codestral, and Maverick are nearly clean.

\paragraph{Truncation and the 16k ablation.}
The 2{,}258 length failures concentrate in Qwen (830, 11.8\% of its
attempts), DeepSeek-Pro (760), DeepSeek-Flash (399), and GLM (260). A fixed
4{,}096-token completion budget is a protocol choice that systematically
truncates long full-document rewrites, so we \emph{re-ran} the length-failed
cells at \code{max\_tokens}${=}16{,}384$ under the same prompt and provider
pinning.
The rerun covered 1{,}947 of 2{,}258 length cells, sampling the
highest-cost route (Qwen: 519 of its 830 cells), 40 cells were lost
to provider rate limits during the rerun. Results
(Table~\ref{tab:ablation}): \textbf{77\% of rerun truncation cells produce a
compiling repair once the cap is lifted} (1{,}504/1{,}947), reaching
\textbf{90\%} for Qwen. Truncation was therefore predominantly a protocol
artifact, not model inability. Folding observed conversions back into
\itt{} accounting, Qwen's full-set \critt{} rises from 56.1\% to at least
62.7\% observed (${\approx}66.7\%$ extrapolating its 90\% sample rate to all
830 cells), and DeepSeek-Pro's from 65.8\% to 73.8\%, while Grok, Codestral,
and Maverick are unaffected (zero or near-zero length cells). Two caveats:
49 GLM cells migrated from \emph{length} to \emph{empty} at 16k (failure
class migration, not repair), and 73 of the new compiles are exact golden
reverts. Model rankings under a 16k protocol would compress but not invert:
Grok's lead persists, but the gap to the truncation-affected models roughly
halves.

\begin{table}[t]
\caption{16k truncation ablation: former \code{length} cells rerun at
\code{max\_tokens}${=}16{,}384$ (Qwen sampled 519/830).}
\label{tab:ablation}
\scriptsize
\begin{tabular}{@{}lrrrr@{}}
\toprule
Model & Rerun & API-ok & Compile & Conv.\ rate \\
\midrule
qwen/qwen3.7-max & 519 & 517 & 467 & \textbf{90.0\%} \\
deepseek/deepseek-v4-pro & 760 & 757 & 559 & 73.6\% \\
z-ai/glm-5.2 & 260 & 210 & 190 & 73.1\% \\
deepseek/deepseek-v4-flash & 399 & 360 & 285 & 71.4\% \\
mistralai/codestral-2508 & 9 & 5 & 3 & 33.3\% \\
\midrule
All & 1947 & 1849 & 1504 & 77.2\% \\
\bottomrule
\end{tabular}\\[1pt]
{\footnotesize Conv.\ rate $=$ compile$/$rerun. 39 DS-Flash and 1 GLM cells
lost to rate limits, 4 Codestral cells still truncate at 16k.}
\end{table}

\paragraph{Equivalent mutants / golden reverts.}
1{,}527 API-ok candidates (3.7\%) match the golden SHA-256 exactly, ranging
from 0.9\% (Codestral) to 7.7\% (GLM). Reversion compiles but is a weak
repair signal: it solves the instance by undoing the operator. Candidate
explanations (edit style, delivery selection, training familiarity) are
observationally indistinguishable here. The restoration oracle below provides
the complementary continuous view.

\section{Results: Restoration Oracle}
\label{sec:restoration}

Compile success is necessary but not sufficient: a candidate can compile
while deleting or rewriting content~\cite{smith2015overfitting,texfixv03}. We
therefore score every compiling, non-revert candidate with a restoration
oracle: compile the candidate, extract text (\code{pdftotext}), and compute
token-LCS Dice similarity \spdf{} against the golden's extracted text.

\paragraph{Scorability accounting.}
Of 32{,}937 candidates, 4{,}708 rows (14.3\%) are \emph{unscorable}: their
golden compiles to a valid but textless PDF. These concentrate in Typst
(4{,}435 rows from template/library seeds whose top-level file defines
functions without rendering content) plus graphics-only LaTeX (259).
Restoration similarity is undefined, not zero, for such instances, so they
are excluded with this explicit accounting. A further 100 candidates
failed re-compilation. The scored set is \textbf{28{,}129} candidates.
Markdown ``restoration'' compares pandoc HTML output rather than PDF text
and is reported separately.

\subsection{RQ6: Do compiling repairs preserve content?}
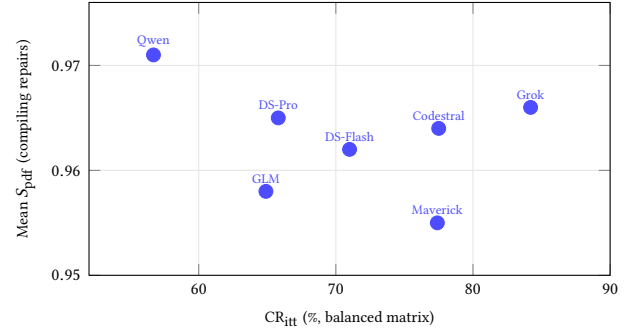
\begin{figure}[t]
\centering
\begin{tikzpicture}
\begin{axis}[
  width=\columnwidth,
  height=5.2cm,
  xlabel={\critt{} (\%, balanced matrix)},
  ylabel={Mean \spdf{} (compiling repairs)},
  xmin=52, xmax=90,
  ymin=0.950, ymax=0.976,
  grid=major,
  grid style={gray!20},
  tick label style={font=\scriptsize},
  label style={font=\scriptsize},
  nodes near coords,
  point meta=explicit symbolic,
  every node near coord/.append style={font=\tiny,anchor=south},
]
\addplot[only marks,mark=*,mark size=2.6pt,blue!70] coordinates {
  (84.2,0.966) [Grok]
  (77.5,0.964) [Codestral]
  (77.4,0.955) [Maverick]
  (71.0,0.962) [DS-Flash]
  (65.8,0.965) [DS-Pro]
  (64.9,0.958) [GLM]
  (56.7,0.971) [Qwen]
};
\end{axis}
\end{tikzpicture}
\caption{Compile rate vs.\ content restoration. The two oracles disagree in
the tails: Qwen (lowest \critt{}) restores best among its successes,
Maverick (third-highest \critt{}) restores worst.}
\label{fig:scatter}
\end{figure}

Not reliably (Figure~\ref{fig:scatter}). Per model, \textbf{13.6--18.5\% of compiling repairs fall below
\spdf{}${=}0.95$}: they lose or alter a material fraction of document text
(Table~\ref{tab:restoration}). Median \spdf{} is 1.0 for every model: most
repairs are faithful, but a heavy tail of content-destroying rewrites
persists everywhere.

\begin{table}[t]
\caption{Restoration among compiling non-revert repairs (goldens with
extractable text, LaTeX+Typst+Markdown rows, means over
per-candidate \spdf{}).}
\label{tab:restoration}
\scriptsize
\begin{tabular}{@{}lrrr@{}}
\toprule
Model & $n$ & Mean \spdf{} & $<$0.95 share \\
\midrule
qwen/qwen3.7-max & 3336 & \textbf{0.971} & \textbf{13.9\%} \\
x-ai/grok-4.3 & 4619 & 0.966 & 13.6\% \\
deepseek/deepseek-v4-pro & 3717 & 0.965 & 13.9\% \\
mistralai/codestral-2508 & 4581 & 0.964 & 16.1\% \\
deepseek/deepseek-v4-flash & 3853 & 0.962 & 16.4\% \\
z-ai/glm-5.2 & 3644 & 0.958 & 14.4\% \\
meta-llama/llama-4-maverick & 4379 & 0.955 & 18.5\% \\
\bottomrule
\end{tabular}
\end{table}

\paragraph{Restoration rank diverges from compile rank.}
Qwen3.7-max, \emph{last} under \critt{}, has the \emph{best} restoration
among its compiling repairs. Llama-4 Maverick, third under \critt{}, has
the worst (18.5\% below 0.95). Grok combines the top compile rate with
near-top restoration. The ranking is robust to the threshold choice: at
$\tau{=}0.90/0.95/0.99$ the per-model below-threshold shares span
8.1--11.9\%, 13.6--18.5\%, and 26.9--35.5\% respectively, with Maverick
worst and the Grok/Qwen/DeepSeek-Pro cluster best at every $\tau$.
The divergence is modest in means (0.955--0.971) but
systematic in tails, and it inverts for specific models: optimizing
leaderboard position under a compile-only oracle rewards exactly the
aggressive full-rewrite behavior that the restoration oracle penalizes.

\paragraph{Per-format and per-tier.}
Mean \spdf{}: LaTeX 0.975 ($n{=}21{,}300$), Typst 0.970 ($n{=}4{,}328$),
Markdown HTML-similarity 0.847 ($n{=}2{,}501$, a different construct: models reformat Markdown freely even when conversion succeeds). Tier~3 shows the
worst restoration tail (13.5\% below 0.95 vs.\ 10.4\% for Tier~1),
consistent with harder faults inducing larger rewrites.

\section{Discussion}
\label{sec:discussion}

\paragraph{Fault models change conclusions.}
The validation results (Section~\ref{sec:validation}) quantify what
taxonomy-grounded injection buys: 5.6--9.2\,pp harder compile targets and
substantially harder exact restoration, replicated across three model
families, with real human faults harder still. Benchmarks built from brace
typos alone will overstate field performance.

\paragraph{Two oracles, one leaderboard.}
Compile-only scoring and restoration scoring disagree about models in the
tails. End-to-end document-repair products should report both: \critt{} for
deployment reality (including serving-stack delivery), \crapi{} for model
skill, and a restoration statistic for content fidelity.

\paragraph{Delivery is a first-order outcome, and attributable.}
With provider-pinned routing and a released provider ledger, the Qwen and
DeepSeek-Pro delivery deficits are attributable to first-party serving
stacks. Cost intensity is likewise first-order: recorded provider cost per
1{,}000 attempts ranges from ${\approx}$\$0.45 (DS-Flash) and \$0.79
(Maverick) to \$4.98 (Grok) and \textbf{\$14.74} (Qwen). The most expensive
model in the panel yields the weakest \itt{} rate, a caution for
cost-unaware leaderboards.

\paragraph{Typst gap.}
The Typst deficit appears across most models and is a useful signal of
training-data currency for newer languages.

\paragraph{Matrix completeness.}
Unevaluated schedule cells are disclosed, not hidden: the balanced
6{,}613$\times$7 matrix removes any unevenness concern for primary claims,
and released checkpoints allow anyone to evaluate the remaining freeze
instances under the identical protocol at proportional cost.

\section{Threats to Validity}
\label{sec:threats}

\paragraph{Construct validity.}
Compile success is not semantic or visual correctness. \spdf{} measures
extracted-text similarity, not layout, and its tokenizer does not yet fold
Unicode normalization or ligatures. Tier labels are design choices whose
ordinal meaning varies by format (RQ3). The taxonomy targets localized hard
crashes. Reconfiguration-class faults (29.1\% of mined candidates) are out of
scope by design. The Markdown restoration construct differs from PDF text
similarity. The zero-shot, no-diagnostics protocol is deliberately
conservative: real repair tools have the compiler error available, so the
absolute rates reported here are lower bounds on assisted repair, and a
diagnostics-in-prompt condition is a planned extension.

\paragraph{Internal validity.}
Mutations are synthetic. Residual broken compiles (5.7\%) show soft-track and
Markdown oracles are imperfect. Engine versions are pinned on one machine.
Other TeX Live / Typst versions may differ~\cite{tan2024tex}. Open-coding
$\kappa{=}0.34$ reflects granularity asymmetry (Section~\ref{sec:kappa}). The
operationalizability check does not substitute for a second human codebook
round. Annotator~1 is the author. The truncation ablation reallocates
length-failed cells but cannot rule out residual prompt--model interaction.

\paragraph{External validity.}
Seven OpenRouter routes are not all LLMs. Frontier closed models beyond Grok
are absent. Seeds are open templates and examples, not full multi-file
research monographs with private assets. In addition, 1{,}158 seeds compile
to textless PDFs and are excluded from restoration with explicit
accounting. The mined
taxonomy is LaTeX-only. Typst/Markdown operators are analogues. TeX.SE
dominates the mined corpus (143/168) despite dual sampling. English-language
bias remains. The real-error evaluation is an 88-case study, not a population
sample.

\section{Conclusion}
\label{sec:conclusion}

TeXFix-Bench closes the loop from mined human faults to large-scale LLM
evaluation: an 18-category taxonomy derived from 168 verified hard crashes
with transparent reliability analysis, \docmut{}, a 48-operator multi-format
mutation library whose faults are 5.6--9.2\,pp harder than pattern-based
injection across three model families and bracketed from below by real human
faults (67\%), a 10{,}437-instance benchmark, and a seven-model,
provider-pinned campaign whose primary claims hold on a complete
6{,}613$\times$7 balanced matrix. Under the compile oracle we find a
27.5-point \itt{} spread and a persistent Typst gap. Under the restoration
oracle we find that 13.6--18.5\% of compiling repairs materially alter
document text, with restoration rank diverging from compile rank. Document
repair needs both oracles, and fault models grounded in evidence rather than
convenience. All artifacts are released.

\begin{acks}
Independent second-annotator coding is gratefully acknowledged.
\end{acks}

\section*{Data Availability}
\docmut{} v0.2.0 (MIT): \url{https://github.com/prajwal-svm/docmut}. The
taxonomy annotation data (open codes, synonym handbook, $\kappa$ audit
trail), campaign ledgers (including the per-request provider ledger),
engine-gate and restoration results, cross-benchmark sets, and harness
scripts are archived in a companion Zenodo research package
(\url{https://doi.org/10.5281/zenodo.21831797}). Provider API keys and Authorization headers are excluded.

\bibliographystyle{ACM-Reference-Format}
\bibliography{references}

\appendix
\section{Operator Codes Observed in the Freeze}
\label{app:ops}
Table~\ref{tab:ops-all} lists operator codes with nonzero instance counts in
the v0.4 freeze. Full rationales ship with the \docmut{} catalog.

\begin{table}[t]
\caption{Operators with support in the v0.4 instance freeze.}
\label{tab:ops-all}
\scriptsize
\begin{tabular}{@{}lr|lr@{}}
\toprule
Code & $n$ & Code & $n$ \\
\midrule
TEX-BRC-DRP & 841 & TYP-CTB-UNC & 587 \\
TEX-BRC-STR & 679 & TYP-FNC-UNC & 671 \\
TEX-ENV-UNC & 671 & TYP-STR-UNC & 666 \\
TEX-ENV-REN & 675 & TYP-TPE-WRG & 458 \\
TEX-CMD-TRP & 730 & TYP-VAR-UDF & 187 \\
TEX-ARG-DRP & 669 & TYP-DCT-DRP & 154 \\
TEX-ENV-SWP & 43 & TYP-REF-UDF & 91 \\
TEX-ITM-MSN & 344 & TYP-IMP-DRP & 86 \\
TEX-CLS-DRP & 393 & TYP-MTH-UNC & 90 \\
TEX-PKG-DRP & 252 & TYP-PGE-SZE & 183 \\
TEX-MTH-DLR & 334 & TYP-SET-INV & 75 \\
TEX-SHL-ESC & 391 & MD-HTML-UNC & 230 \\
TEX-GLS-UDF & 393 & MD-IMG-BRK & 160 \\
TEX-HYP-DRV & 81 & MD-HDR-MLF & 146 \\
TEX-MTH-DSP & 33 & MD-CDE-UNC & 33 \\
TEX-LVL-SFT & 12 & MD-TBL-MLF & 28 \\
TEX-MTH-REL & 13 & MD-LNK-BRK & 8 \\
TEX-LBL-DUP & 13 & MD-YML-BRK & 1 \\
\bottomrule
\end{tabular}
\end{table}

\end{document}